\documentstyle[emnlp]{article}

\newcommand{\np}{$\rm [_{NP}$}
\newcommand{\vp}{$\rm [_{VP}$}
\newcommand{\pp}{$\rm [_{PP}$}

\title{Attaching Multiple Prepositional Phrases: Generalized Backed-off
Estimation \\ ~~}

\author{ Paola Merlo \\
	IRCS-U. of Pennsylvania\\
        LATL-University of Geneva \\ 
        2 rue de Candolle \\
        1204 Geneva, Switzerland \\
        {\tt merlo@latl.unige.ch}\And
	Matthew W. Crocker \\
        Centre for Cognitive Science \\ 
        University of Edinburgh \\ 
        2 Buccleuch Place, \\
        Edinburgh, UK EH8 9LW\\{\tt mwc@cogsci.ed.ac.uk}\And
        Cathy Berthouzoz \\
        LATL \\
	University of Geneva \\ 
        2 rue de Candolle \\
        1204 Geneva, Switzerland \\
        {\tt berthouzoz@latl.unige.ch}
}
\begin{document}
\maketitle
\bibliographystyle{acl}

\begin{abstract}

There has recently been considerable interest in the use of lexically-based
statistical techniques to resolve prepositional phrase attachments. To our
knowledge, however, these investigations have only considered the problem of
attaching the first PP, i.e., in a [V NP PP] configuration.  In this paper, we
consider one technique which has been successfully applied to this problem,
backed-off estimation, and demonstrate how it can be extended to deal with the
problem of multiple PP attachment. The multiple PP attachment introduces two
related problems: sparser data (since multiple PPs are naturally rarer), and
greater syntactic ambiguity (more attachment configurations which must be
distinguished). We present and algorithm which solves this problem through
re-use of the relatively rich data obtained from first PP training, in
resolving subsequent PP attachments.
\end{abstract}

\section{Introduction}

Ambiguity is the most specific feature of natural languages, which
sets them aside from programming languages, and which is at the root
of the difficulty of the parsing enterprise, pervading languages at
all levels: lexical, morphological, syntactic, semantic and
pragmatic. Unless clever techniques are developed to deal with
ambiguity, the number of possible parses for an average sentence (20
words) is simply intractable. In the case of prepositional phrases,
the expansion of the number of possible analysis is the Catalan number
series, thus the number of possible analyses grows with a function
that is exponential in the number of Prepositional Phrase
\cite{church-patil82}.  One of the most interesting topics of debate
at the moment, is the use of frequency information for automatic
syntactic disambiguation.  As argued in many pieces of work in the AI
tradition \cite{marcus80,crain-steedman85,altmann-steedman88,hirst87},
the exact solution of the disambiguation problem requires complex
reasoning and high level syntactic and semantic knowledge. However,
current work in part-of-speech tagging has succeeded in showing that
it is possible to carve one particular subproblem and solve it {\em by
approximation\/} --- using statistical techniques --- independently
of the other levels of computation.

In this paper we consider the problem of prepositional phrase (PP)
ambiguity. While there have been a number of recent studies concerning
the use of statistical techniques for resolving single PP attachments,
i.e. in constructions of the form [V NP PP], we are unaware of
published work which applies these techniques to the more general, and
pathological, problem of multiple PPs, e.g. [V NP PP1 PP2 ...]. In
particular, the multiple PP attachment problem results in sparser
data which must be used to resolve greater ambiguity: a strong test
for any probabilistic approach.

We begin with an overview of techniques which have been used for PP
attachment disambiguation, and then consider how one of the most
successful of these, the backed-off estimation technique, can be applied
to the general problem of multiple PP attachment.

\section{Existing Models of Attachment}

Attempts to resolve the problem of PP attachment in computational linguistics
are numerous, but the problem is hard and success rate typically depends on the
domain of application. Historically, the shift from attempts to resolve the
problem completely, by using heuristics developed using typical AI techniques
\cite{jensen-binot87,marcus80,crain-steedman85,altmann-steedman88} has left the
place for attempts to solve the problem by less expensive means, even if only
approximately.  As shown by many psycholinguistic and practical studies
\cite{ford-ea82,taraban-mcclelland88,whittemore-ea90}, lexical information is
one of the main cues to PP attachment disambiguation.

In one of the earliest attempts to resolve the problem of PP attachment
ambiguity using lexical measures, Hindle and Rooth \shortcite{hindle-rooth93}
show that a measure of mutual information limited to lexical association can
correctly resolve 80\% of the cases of PP attachment ambiguity, confirming the
initial hypothesis that lexical information, in particular co-occurrence
frequency, is central in determining the choice of attachment.

The same conclusion is reached by Brill and Resnik \shortcite{brill-resnik94}.
They apply transformation-based learning \cite{brill93} to the problem of
learning different patterns of PP attachment. After acquiring 471 patterns of
PP attachment, the parser can correctly resolve approximately 80\% of the
ambiguity. If word classes \cite{resnik93} are taken into account, only 266
rules are needed to perform at 80\% accuracy.

Magerman and Marcus \shortcite{magerman-marcus91} report 54/55 correct PP
attachments for Pearl, a probabilistic chart parser, with Earley style
prediction, that integrates lexical co-occurrence knowledge into a probabilistic
context-free grammar.  The probabilities of the rules are conditioned on the
parent rule and on the trigram centered at the first input symbol that would be
covered by the rule. Even if the parser has been tested only in the direction
giving domain, where the behaviour of prepositions is very consistent, it shows
that a mixture of lexical and structural information is needed to solve the
problem successfully.

Collins and Brooks \shortcite{collins-brooks} propose a 4-gram model for PP
disambiguation which exploits backed-off estimation to smooth null events (see
next section). Their model achieves 84.5\% accuracy. The authors point out that
prepositions are the most informative element in the tuple, and that taking low
frequency events into account improves performance by several percentage
points. In other words, in solving the PP attachment problem, backing-off is
not advantageous unless the tuple that is being tested is not present in the
training set (it has zero counts). Moreover, tuples that contain prepositions
are the most informative.

The second result is roughly confirmed by Brill and Resnik, (ignoring
the importance of n2 when it is a temporal modifier, such as {\em
yesterday, today}). In their work, the top 20 transformations learned
are primarily based on specific prepositions.

\section{Back-off Estimation}

The PP attachment model presented by Collins and Brooks
\shortcite{collins-brooks} determines the most likely attachment for a
particular prepositional phrase by estimating the probability of the
attachment. We let $C$ represent the attachment event, where $C=1$
indicates that the PP attaches to the verb, and $C=2$ indicates
attachment to the object NP. The attachment is conditioned by the
relevant head words, a 4-gram, of the VP.

\begin{itemize}

\item Tuple format: ($C$, v, n1, p, n2)
\item So: {\em John read [[the article] [about the budget]]}
\item Is encoded as: (2, read, article, about, budget)\\
\end{itemize}

Using a simple maximal likelihood approach, the best attachment for a
particular input tuple (v,n1,p,n2) can now be determined from the
training data via the following equation:

\begin{equation}
argmax_i\;\hat{p}(C=i|v, n1, p, n2)=\frac{f(i,v,n1,p,n2)}{f(v,n1,p,n2)}
\end{equation}

Here $f$ denotes the frequency with which a particular tuple
occurs. Thus, we can estimate the probability for each configuration
$1\le i\le 2$, by counting the number of times the four head words
were observed in that configuration, and dividing it by the total
number of times the 4-tuple appeared in the training set.

While the above equation is perfectly valid in theory, sparse data
means it is rather less useful in practice. That is, for a particular
sentence containing a PP attachment ambiguity, it is very likely
that we will never have seen the precise (v,n1,p,n2) quadruple before
in the training data, or that we will have only seen it
rarely.\footnote{Though as Collins and Brooks point out, this is less
of an issue since even low counts are still useful.} To address this
problem, they employ backed-off estimation when zero counts occur
in the training data. Thus if $f(v,n1,p,n2)$ is zero, they `back-off'
to an alternative estimation of $\hat{p}$ which relies on 3-tuples
rather than 4-tuples:

\begin{eqnarray}
\hat{p}_3(C=i|v, n1, p, n2)  = & &\\ 
\frac{f(i,v,n1,p)+f(i,v,p,n2)+f(i,n1,p,n2)}{f(v,n1,p)+f(v,p,n2)+f(n1,p,n2)} & &\nonumber
\end{eqnarray}

Similarly, if no 3-tuples exist in the training data, they back-off further:

\begin{eqnarray}
\hat{p}_2(C=i|v, n1, p, n2) =& &\\ 
	 \frac{f(i,v,p)+f(i,n1,p)+f(i,p,n2)}{f(v,p)+f(n1,p)+f(p,n2)} & &\nonumber
\end{eqnarray}

\begin{equation}
\hat{p}_1(C=i|v, n1, p, n2) = \\ \frac{f(i,p)}{f(p)}
\end{equation}

The above equations incorporate the proposal by Collins and Brooks
that only tuples including the preposition should be considered,
following their results that the preposition is the most informative
lexical item. Using this technique, Collins and Brooks achieve an
overall accuracy of 84.5\%.

\section{The Multiple PP Attachment Problem}

Previous work has focussed on the problem of single PP attachment, in
configurations of the form [V NP PP] where both the NP and the PP are assumed
to be attached within the VP. The algorithm presented in the previous section,
for example, simply determines the maximally likely attachment event (to NP or
VP) based on the supervised training provided by a parsed corpus. The broader
value of this approach, however, remains suspect until it can be demonstrated
to apply more generally. We now consider how this approach -- and the use of
lexical statistics in general -- might be naturally extended to handle the more
difficult problem of multiple PP attachment. In particular, we investigate the
PP attachment problem in cases containing two PPs, [V NP PP1 PP2], and three
PPs, [V NP PP1 PP2 PP3], with a view to determining whether n-gram based parse
disambiguation models which use the backed-off estimate can be usefully
applied. Multiple PP attachment presents two challenges to the approach:

\begin{enumerate}

\item For a single PP, the model must make a choice between two
structures. For multiple PPs, the space of possible structural configurations
 increases
dramatically, placing increased demands on the disambiguation
technique.
\item Multiple PP structures are less frequent, and contain more words,
than single PP structures.  This substantially increases the sparse data
problems when compared with the single PP attachment case.
\end{enumerate}

\subsection{Materials and Method}

To carry out the investigation, training and test data were obtained
from the Penn Tree-bank, using the {\tt tgrep} tools to extract tuples
for 1-PP, 2-PP, and 3-PP cases. For the single PP study, VP attachment
was coded as 1 and  NP attachment was coded as 2. A database of
quadruples of the form {\it (configuration,v,n,p)\/} was then created. The
table below shows the two configurations and their frequencies in the
corpus.

\vspace{0.3cm}
\begin{center}
\begin{tabular}{||l|l|r||}\hline \hline 
Configuration 		&  Structure			& Counts\\ \hline
 1& \vp NP PP ]		&  7740\\
 2& \vp \np PP ]]	& 12223\\ \hline \hline
\end{tabular}
\end{center}
\vspace{0.3cm}

The same procedure was used to create a database of 6-tuples
{\it (configuration,v,n1,p1,n2,p2) \/} for the attachment of 2 PPs.  The values for the
configuration varies over a range 1..5, corresponding to the 5 grammatical
structures possible for 2 PPs, shown and exemplified below with their counts in
the corpus.\footnote{We did not consider the left-recursive NP structure for
the 2 PP (or indeed 3 PP) cases. Checking the frequency of their occurrences
revealed that there were only 2 occurrences of \vp \np \np \np PP] PP]]]
structures in the corpus.}
\vspace{0.3cm}

\begin{center}
\begin{tabular}{||l|l|r||}\hline \hline
Config &  Structure			& Counts\\ \hline
 1& \vp V NP PP PP]			&  535\\
 2& \vp V \np NP PP] PP]		& 1160\\
 3& \vp V \np \pp P \np NP PP ]]]]	& 1394\\
 4& \vp V NP \pp  \np NP PP]]]		& 1055\\
 5& \vp V \np NP PP PP]]		& 539  \\ \hline \hline 
\end{tabular}
\end{center}
\vspace{0.3cm}

\begin{enumerate}
\item  The agency said it  will {\bf keep} the {\bf debt under review for} possible further downgrade.
\item Penney decided to {\bf extend} its {\bf involvement with} the {\bf service for} at least five years.
\item The bill was then sent back to the House to resolve the question of how
to {\bf address} budget {\bf limits on} credit {\bf allocations for} the Federal Housing Administration.
\item  Sears officials insist they don't intend to {\bf abandon\/} the everyday
pricing {\bf approach in} the {\bf face of} the poor results.
\item Mr. Ridley hinted at this motive in {\bf answering questions from}
members of {\bf Parliament after} his announcement 
\end{enumerate}
\vspace{0.3cm}

Finally, a database of 8-tuples {\em (configuration,v,n1,p1,n2,p2,n3,p3)\/} was
created for 3 PPs. The value of the configuration varies over a range 1..14,
corresponding to the 14 structures possible for 3 PPs, shown in Table
\ref{table14configs} with their counts in the corpus.

\begin{table*}
\begin{center}
\begin{tabular}{||l|l|r||}\hline \hline 
Configuration &  Structure			& Counts\\ \hline
 1&  \vp   V NP PP PP PP ]			& 15\\
 2&  \vp   V NP PP \pp P \np PP ]]]		& 86\\
 3&  \vp   V \np NP PP] PP PP ]			& 63\\
 4&  \vp   V \np NP PP] \pp p \np PP ]]]		& 168\\
 5&  \vp   V \np \pp P \np NP PP ]]] PP]	& 81\\
 6&  \vp   V \np \pp P \np NP PP ]]PP ]]	& 31\\
 7&  \vp   V \np \pp P \np NP PP PP ]]]]	& 47\\
 8&  \vp   V \np \pp P \np NP \pp P \np PP ]]]]]]	&142\\
 9&  \vp   V NP \pp  \np NP PP]] PP ]		& 47\\
 10& \vp   V NP \pp  \np NP PP  PP ]]]		& 34\\
 11& \vp   V NP \pp  \np NP \pp P \np PP ]]]]]	&80\\
 12& \vp   V \np NP PP PP] PP ]			&20\\  
 13& \vp   V \np NP PP PP PP ]]			&21\\  
 14& \vp   V \np NP PP  \pp P \np PP ]]]]	&72\\  \hline \hline 
\end{tabular}
\end{center}
\caption{Corpus counts for the 14 structures possible for 3-PP sequences.}
\label{table14configs}
\end{table*}

The above datasets were then split into training and test sets by automatically
extracting stratified samples. For PP1, we extracted quadruples of about 5\% of
the total (1014/19963). We then created a test set for PP2 which is a subset of
the PP1 test set, and approximately 10\% of the 2 PP tuples
(464/4683). Similarly, the test set for PP3 is a subset of the PP2 test set of
approximately 10\% (94/907). It is important that the test sets are subsets to
ensure that, e.g., a PP2 test case doesn't appear in the PP1 training set,
since the PP1 data is used by our algorithm to estimate PP2 attachment, and
similarly for the PP3 test set.

\subsection{Does Distance Matter?}

In exploring multiple PP attachment, it seems natural to investigate the
effects of the distance of the PP from the verb. The following table reports
accuracy of noun-attachment, when the attachment decision is conditioned only
on the preposition and on the distance -- in other words, when estimating
$\hat{p}(1|p,d)$ where 1 is the coding of the attachment to the noun, $p$ is
the preposition and $d = \{1,2,3\}$.\footnote{These figures are to be taken
only as an indication of a trend, as they represent the accuracy obtained by
testing on the training data.  Moreover, we are only considering 2 attachment
possibilities for each preposition, either it attaches to the verb or it
attaches to the lowest noun.}

\vspace{0.3cm}
\begin{center}
\begin{tabular}{||l|r|r|r|r|r||} \hline \hline 
	& 1 PP	& 2 PP	& 3 PP	& Total	& All 	\\ \hline
Count	& 20299	& 4711	& 939	& 25949 & 25949 \\ \hline
Correct & 15173	& 3525  & 755   & 19453 & 19349 \\ \hline
\%  & 74.7  & 74.8  & 80.4  & 75    & 74.5 \\ \hline \hline
\end{tabular}
\end{center}
\vspace{0.3cm}

It can be seen from these figures that conditioning the attachment according to
both preposition and distance results in only a minor improvement in
performance, mostly because separating the biases according to preposition {\em
distance} increases the sparse data problem.  It must be noted, however, that
counts show a steady increase in the proportion of low attachments for PP
further from the verb, as shown in the table below. The simplest explanation
of this fact is that more (inherently) noun-attaching prepositions must be
occurring in 2nd and 3rd positions. This predicts that the distribution of
preposition occurrences changes from PP1 to PP3, with an increase in the
proportion of low attaching PPs.  

\vspace{0.3cm}
\begin{center}
\begin{tabular}{||l|c|c|c|c||}\hline \hline
	& 1 PP	& 2 PP	& 3 PP	& Total \\ \hline 
Count	& 20299	& 4711	& 939	& 25949 \\
Low	& 12223	& 3063	& 706	& 15992 \\		
\% Low	& 60.2	& 65.0	& 75.1	& 61.6 \\ \hline \hline 
\end{tabular}
\end{center}
\vspace{0.3cm}

Having established that the distance parameter is not as influential a factor
as we hypothesized, we exploit the observation that attachment
preferences do not significantly change depending on the distance of the PP
from the verb.  In the following section, we discuss an extension of the
back-off estimation model that capitalizes on this property.

\section{The Generalized Backed-Off  Algorithm}

The algorithm for attaching the first preposition is almost identical to that
of Collins and Brooks \shortcite{collins-brooks}, and we follow them in
including only tuples which contain the preposition. We do not, however, use
the final noun (following the preposition) in any of our tuples, thus basing
our model of PP1 on three, rather than four, head words.

\vspace{0.3cm}
{\bf Procedure B1:} 
\vspace{0.3cm}

The most likely configuration is: \\

$arg\;max_{i}\;\hat{p}_1(C_2=i|v,n,p)$, where $1\le i\le2$

\begin{enumerate}

\item IF $f(v,n,p)>0$ THEN \\
             $\hat{p}_1(i|v,n,p)=\frac{f(i,v,n,p)}{f(v,n,p)}$
\item ELSEIF $f(v,p)+f(n,p)>0$ THEN \\
             $\hat{p}_1(i|v,n,p)=\frac{f(i,v,p)+f(i,n,p)}
                                        {f(v,p)+f(n,p)}$
\item ELSEIF $f(p)>0$ THEN \\$\hat{p}_1(i|v,n,p)=\frac{f(i,p)}{f(p)}$
\item ELSE $\hat{p}_1(1|v,n,p)=0, \hat{p}_1(2|v,n,p)=1$
\end{enumerate}

In this case $i$ denotes the attachment configuration: $i=1$ is VP
attachment, $i=2$ is NP attachment. The subscript on $C_2$ is used
simply to make clear that $C$ has 2 possible values. In the subsequent
algorithms, $C_5$ and $C_{14}$ are used to indicate the larger sets of
configurations.

 The algorithm used to handle the cases containing 2PPs is shown in Figure
\ref{procedureb2}, where $j$ ranges over the five possible attachment
configurations outlined above.

\vspace{0.3cm}
\begin{figure*}
{\bf Procedure B2} 
\vspace{0.3cm}

The most likely configuration is: \\
	$arg\;max_{j}\;\hat{p}_2(C=j|v,n1,p1,n2,p2)$, 
where  $1\le j\le5$

\begin{tabbing}
1. IF $f(v,n1,p1,n2,p2)>0$ THEN  \= \\
		\>$\hat{p}_2(j)=\frac{f(j,v,n1,p1,n2,p2)}{f(v,n1,p1,n2,p2)}$ \\
\\
2. ELSEIF  $f(n1,p1,n2,p2)+f(v,p1,n2,p2)+f(v,n1,p1,p2)>0$ THEN \\
\>	$\hat{p}_2(j)=\frac{f(j,n1,p1,n2,p2)+f(j,v,p1,n2,p2)+f(j,v,n1,p1,p2)}
                           {f(n1,p1,n2,p2)+f(v,p1,n2,p2)+f(v,n1,p1,p2)}$\\
\\
3. ELSEIF $f(p1,n2,p2)+f(v,p1,p2)+f(n1,p1,p2)>0$ THEN \\
\>	$\hat{p}_2(j)=\frac{f(j,p1,n2,p2)+f(j,v,p1,p2)+f(j,n1,p1,p2)}
                           {f(p1,n2,p2)+f(v,p1,p2)+f(n1,p1,p2)}$ \\
\\
4. ELSE {\bf Competitive Backed-off Estimate}
\end{tabbing}
\rule{\textwidth}{0.05cm}
\caption{Procedure B2}
\label{procedureb2}
\end{figure*}

The first three steps use the standard backed-off estimation, again including
only those tuples containing {\em both} prepositions. However, after
backing-off to three elements, we abandon the standard backed-off estimation
technique. The combination of sparse data, and too few lexical heads, renders
backed-off estimation ineffective. Rather, we propose a technique which makes
use of the richer data available from the PP1 training set. Our hypothesis is
that this information will be useful in determining the attachments of
subsequent PPs as well. This is motivated by our observations, reported in the
previous section, that the distribution of high-low attachments for specific
prepositions did not vary significantly for PPs further from the verb.  The
{\bf Competitive Backed-Off Estimate} procedure, presented below, operates by
initially fixing the configuration of the first preposition (to either the VP
or the direct object NP), and then considers how the second preposition would
be optimally attached into the configuration.

\vspace{0.3cm}
\noindent
{\bf Procedure Competitive Backed-off Estimate} \\
\vspace{0.3cm}
\begin{enumerate}

\item $C'_2$ is the most likely configuration for PP1, \\
      $arg\;max_i\;\hat{p}_1(C'_2=i|v,n1,p1)$
\item $C''_2$ is the preferred configuration for PP2 w.r.t n2, \\
      $arg\;max_i\;\hat{p}_1(C''_2=i|v,n2,p2)$
\item $C'''_2$ is the preferred configuration for PP2 w.r.t n1, \\
      $arg\;max_i\;\hat{p}_1(C'''_2=i|v,n1,p2)$
\item {\bf Find Best Configuration}
\end{enumerate}

\vspace{0.3cm}

First we determine $C'_2$, on which depends the attachment of
p1. We then determine $C''_2$, which indicates the preference for p2 to
attach to the VP or to n2, and $C'''_2$, which is the preference for p2
to attach to the VP or to n1. Given the preferred configurations $C'_2$,
$C''_2$, and $C'''_2$, we now must determine the best of the five
possible configurations, $C_5$, for the entire VP.  

\vspace{0.3cm}
{\bf Procedure Find Best Configuration}
\vspace{0.3cm}
\begin{enumerate}

\item IF $C'_2=1$ and $C''_2=1$ THEN \\ $C_5 \leftarrow 1$
\item ELSEIF $C'_2=1$ and $C''_2=2$ THEN \\ $C_5 \leftarrow 4$
\item ELSEIF $C'_2=2$ and $C''_2=1$ and $C'''_2=1$ THEN \\
	$C_5 \leftarrow 2$
\item ELSEIF $C'_2=2$ and $C''_2=2$ and $C'''_2=1$ THEN \\
	$C_5 \leftarrow  3 $
\item ELSEIF $C'_2=2$ and $C''_2=1$ and $C'''_2=2$ THEN \\
	$C_5 \leftarrow 2 $
\item ELSEIF $C'_2=2$ and $C''_2=2$ and $C'''_2=2$ THEN {\em tie-break} 

\begin{enumerate}

\item IF $f(2,v,n2,p2) < f(2,v,n1,p2)$ THEN \\
	$ C_5 \leftarrow 5$
\item ELSE $C_5 \leftarrow 3$
\end{enumerate}
\end{enumerate}

The tests 1 to 5 simply use the attachment values $C'_2$, $C''_2$, and
$C'''_2$ to determine $C_5$: the best configuration. In the final
instance, step 6, where the $C''_2$ indicates a preference for n2
attachment, and $C'''_2$ indicates a preference for n1 attachment a
tie-break is necessary to determine which noun to attach to. As a
first approximation, we use the frequency of occurrence used in
determining these preferences, rather than the probability for each
preference. That is, we favour the bias for which there is more
evidence, though whether this is optimal remains an empirical
question. For example, if $C''_2$ is based on 4 observations, and
$C'''_2$ is based on 7, then the $C'''_2$ preference is considered
stronger. 

Having constructed the algorithm to determine the best configuration
for 2 PPs, we can similarly generalize the algorithm to handle three.
In this case $k$ denotes one of fourteen possible attachment
configurations shown earlier. The pseudo code for procedure B3 is shown
below, simplified for reasons of space. 

\vspace{0.3cm}
{\bf Procedure B3} 
\vspace{0.3cm}

The most likely configuration is: \\

\noindent
$arg\;max_{k}\;\hat{p}_3(C_{14}=k|v,n1,p1,n2,p2,n3,p3)$, where $1\le k\le14$

\begin{enumerate}

\item IF $f(v,n1,p1,n2,p2,n3,p3)>0)$ THEN \\
        $\hat{p}_3(k)=\frac{f(k,v,n1,p1,n2,p2,n3,p3)}
                             {f(v,n1,p1,n2,p2,n3,p3)}$
\item ELSE Try backing-off to 6 or 5 items \ldots
\item ELSE Competitive Backed-off Estimate:

\begin{enumerate}

\item Use {\bf Procedure B2} to determine $C'_5$, the configuration of p1 and p2
\item Compute $C''_2$, $C'''_2$, $C''''_2$, the preferred attachment of p3 w.r.t n1, n2, n3 respectively
\item Determine the best configuration
\end{enumerate}
\end{enumerate}
\vspace{0.3cm}

Again, we back-off up to two times, always including tuples which
contain the three prepositions. After this, backing-off becomes
unstable, so we use the {\bf Competitive Backed-off Estimate}, as
above, but scaled up to handle the three prepositions and
fourteen possible configurations.

\subsection{Results}

To evaluate the performance of our algorithm, we must first
determine what the expected baseline, or lower-bound on, performance
would be. Given the variation in the number of possible configurations
across the three cases, the performance expected due to chance would
be 50\% for 1 PP, 20\% for 2 PPs, and 7\% for 3 PPs. A better baseline
is the performance that would be expected by simply adopting the most
likely configuration, without regard to lexical items. This is shown
in the table below, with the most frequent configuration
shown in parentheses.

\vspace{0.3cm}
\begin{center}
\begin{tabular}{||l|c|c|c||} \hline \hline
		& PP1(2)	& PP2(5)	& PP3(14) \\ \hline
Total		& 19963		& 4683		& 907 \\
Most Frequent	& 12223(2)	& 1394(3)	& 168(4) \\
Percent Correct	& 61.2\%	& 29.8\%	& 18.5\% \\ \hline  \hline
\end{tabular}
\end{center}
\vspace{0.3cm}

Table \ref{table1} presents the performance of the competitive backed-off
estimation algorithm on the test data. As can be seen, the performance for PP1
replicates the findings of Collins and Brooks, who achieved 84.5\% (using 4
lexical items, compared to our three). For PP2 performance is again high,
recalling that the algorithm is discriminating five possible attachment
configurations, and the baseline expectation was only 29.8\%.  Similarly for
PP3, our performance of 43.6\% accuracy (discriminating fourteen
configurations) far out strips the baseline of 18.5\%.

\begin{table*}
\begin{center}
\begin{tabular}{||l|rr|rr|rr||} \hline \hline
	&\multicolumn{2}{c|}{PP1}&\multicolumn{2}{c|}{PP2}&\multicolumn{2}{c||}{PP3}  \\ \hline
	& Total & Correct & Total & Correct & Total & Correct \\ \hline
No back-off & 300 & 285   & 36    & 35      & 1     & 1 \\
Back-off 1  & 614 & 510	  & 61    & 54      & 1     & 1 \\
Back-off 2  & 100 &  60   & 232   & 161     & 3     & 3 \\
Competitive & NA  &  NA   & 135   & 73      & 89    & 36 \\
Total	    & 1014& 855   & 464   & 323     & 94    & 41 \\ \hline
Percent	    &\multicolumn{2}{c|}{ 84.3\% }   &\multicolumn{2}{c|}{ 69.6\% }&\multicolumn{2}{c||}{43.6\%} \\ \hline \hline
\end{tabular}
\end{center}
\caption{Performance of the competitive backed-off estimation algorithm on the test data.}
\label{table1}
\end{table*}

\section{Conclusions}

The backed-off estimate has been demonstrated to work successfully for
single PP attachment, but the sparse data problem renders it
impractical for use in more complex constructions such as multiple PP
attachment; there are too many configurations, too many head words,
too few training examples. In this paper we have demonstrated,
however, that the relatively rich training data obtained for the first
preposition can be exploited in attaching subsequent PPs. The
algorithm incrementally fixes each preposition into the configuration
and the more informative PP1 training data is exploited to settle the
competition for possible attachments for each subsequent
preposition. Performance is considerably better than both chance and
the naive baseline technique. The generalized backed-off estimation
approach which we have presented constitutes a practical solution to
the problem of multiple PP disambiguation. This further suggests that
backed-off estimation may be successfully integrated into more general
syntactic disambiguation systems.


\section*{Acknowledgments}

We gratefully acknowledge the support of the British Council and the Swiss
National Science Foundation on grant 83BC044708 to the first two authors, and
on grant 12-43283.95 and fellowship 8210-46569 from the Swiss NSF to the first
author.  We thank the audiences at Edinburgh and Pennsylvania for their
useful comments. All errors remain our responsibility.

\end{document}